\documentclass[conference]{IEEEtran}
\ifdefined\pdfminorversion\pdfminorversion=7\fi
\IEEEoverridecommandlockouts

\usepackage{cite}
\usepackage{amsmath,amssymb,amsfonts}
\usepackage{algorithmic}
\usepackage{graphicx}
\usepackage{textcomp}
\usepackage{pifont}
\usepackage{xcolor}
\usepackage{booktabs,tabularx,array}
\usepackage{multirow}
\usepackage{url}
\usepackage[hidelinks]{hyperref}
\def\BibTeX{{\rm B\kern-.05em{\sc i\kern-.025em b}\kern-.08em
    T\kern-.1667em\lower.7ex\hbox{E}\kern-.125emX}}
\begin{document}

\title{AutoRAGBench: A Benchmark for LLM Agents\\Automating RAG System Development}

\author{
\IEEEauthorblockN{Hang Zhang\textsuperscript{1,*}, Yuzhi Pan\textsuperscript{1,*}, Ziyao Zhong\textsuperscript{1,*}, Chaokun Wang\textsuperscript{1}, Shuo Cao\textsuperscript{1}, Zeyu Huang\textsuperscript{1}, Yue Xue\textsuperscript{1},\\Fang Niu\textsuperscript{1}, Bofan Xie\textsuperscript{1}, Guanchen Ge\textsuperscript{1}, Zecheng Li\textsuperscript{1}, Leqi Zheng\textsuperscript{1}, Ziyang Liu\textsuperscript{1}, Wei Wang\textsuperscript{2}, Miao Gu\textsuperscript{2}}
\IEEEauthorblockA{\textsuperscript{1}School of Software, Tsinghua University, Beijing, China\\
\textsuperscript{2}Sina Weibo Corp, Beijing, China\\
\{zhanghang24, panyz24, zhongzy25, caos24, y-xue24, huangzy24, nf21, xbf25, ggc22, zc-li23, zhenglq24,\\ liu-zy21\}@mails.tsinghua.edu.cn, chaokun@tsinghua.edu.cn,
\{robert1, gumiao\}@staff.sina.com}
\thanks{\textsuperscript{*}The authors contribute equally to this work.}
\thanks{This work has been submitted to the IEEE for possible publication.
Copyright may be transferred without notice, after which this version may no longer be accessible.}
}

\maketitle

\begin{abstract}
Retrieval-augmented generation (RAG) is a well-established paradigm for augmenting large language models (LLMs) with information retrieved from external corpora.
However, developing effective RAG systems requires substantial engineering effort and iterative experimentation, as effective design choices vary across tasks and domains.
With rapid advances in LLM agent capabilities, we ask how well these agents can autonomously build RAG systems and how effectively they can improve them through iteration.
In this paper, we propose AutoRAGBench, the first benchmark to evaluate how well LLM agents develop RAG systems from scratch in an open code space.
It comprises nine RAG system development tasks.
For each task, agents build RAG systems from scratch without starter code or prescribed strategies, under a six-hour budget on a single H800 GPU.
They analyze task data, implement RAG pipelines, and iteratively refine their designs based on experimental feedback.
We evaluate nine agent configurations and analyze both the effectiveness of the resulting RAG systems and their development trajectories.
GPT-6.1 Sol with Codex CLI achieves the highest average score of 56.1.
The composite Human-SOTA baseline averages 44.4.
These results show that LLM agents can automate effective RAG system development.
When starting from a human-designed system on each of seven tasks, GPT-6.1 Sol with Codex CLI improves the score by 16.6 points on average while preserving the methods' core mechanisms.
\end{abstract}


\section{Introduction}

\begin{figure*}[t]
    \centering
    \includegraphics[width=\textwidth]{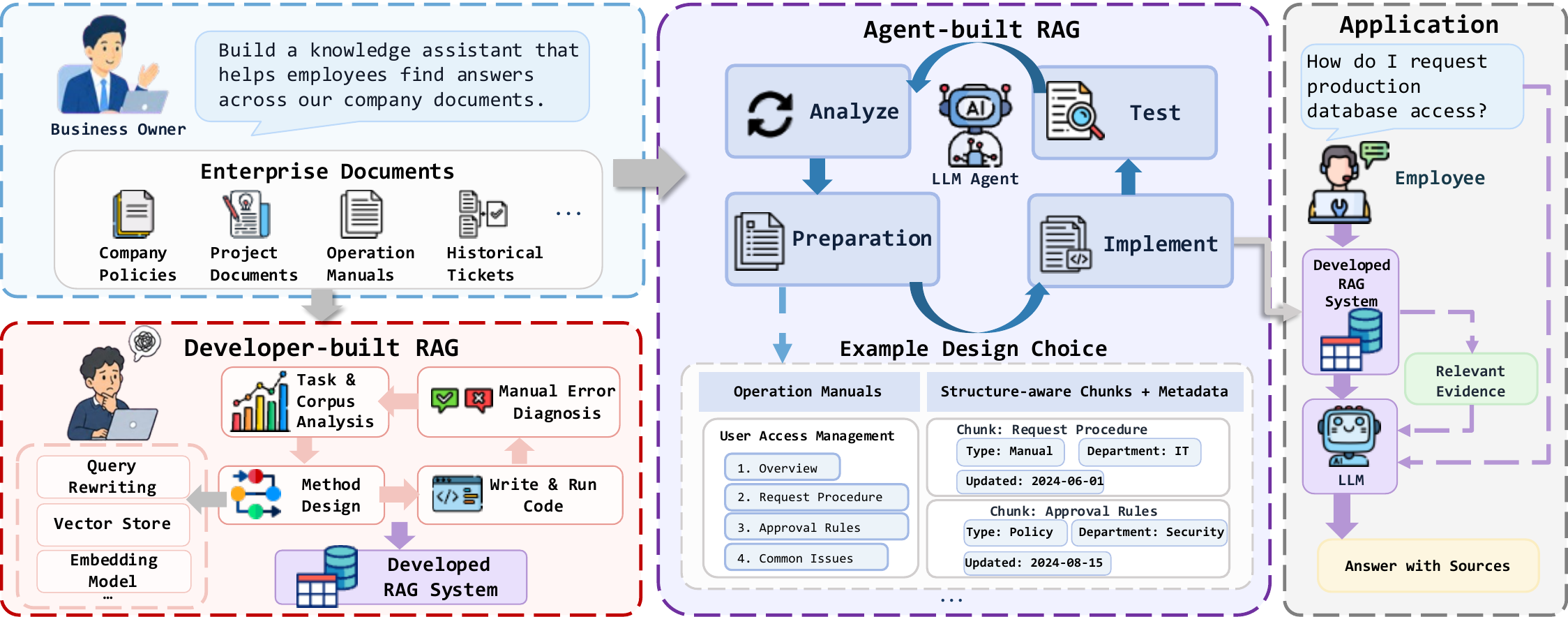}
    \caption{Building a RAG assistant for internal documents. Developers spend considerable time and effort implementing, testing, and refining RAG methods, whereas an LLM agent can efficiently refine the system over successive iterations.}
    \label{fig:motivating-example}
\end{figure*}

Retrieval-augmented generation (RAG)~\cite{huang2026survey,gao2023retrieval,fan2024survey} is a well-established paradigm that augments large language models (LLMs) with information retrieved from external corpora. Retrieved evidence grounds answers in traceable sources, and RAG also supports updates to the external corpus~\cite{wu2026retrieval}. However, accurate and efficient RAG requires task-specific design choices, such as how to process the corpus, build the index, retrieve information, and select evidence~\cite{gao2026modular}. Multi-hop questions\cite{Yang2018HotpotQAAD,Trivedi2021MM} require combining evidence across documents, whereas knowledge-graph questions~\cite{Yih2016TheVO} require entity linking and graph traversal. Developers must therefore choose RAG methods suited to the corpus and the questions users ask, then iteratively improve the system based on test results.

As shown in Fig.~\ref{fig:motivating-example}, a business owner wants an intelligent question-answering (QA) assistant for employees who need information from internal documents.
A RAG system can power this assistant by retrieving passages relevant to each question and supplying them to the LLM as evidence for its answer~\cite{oche2025systematic}.
Manually building an effective RAG system is difficult and time-consuming because developers must analyze the documents, select and implement a retrieval strategy, evaluate its performance on example questions, and refine it based on the results.
Recent advances in LLM agents~\cite{yang2024swe,choi2026longhorizon} have enabled them to automate much of this development cycle.
These capabilities motivate our research question.
How well can LLM agents autonomously build RAG systems across tasks, and how effectively can they improve these systems through iteration?

Existing studies do not directly address this question. First, prior work on RAG evaluation~\cite{rau2024bergen,11629265,NEURIPS2024_1435d2d0,friel2024ragbench} examines how individual components and workflow configurations affect system performance and whether the resulting answers are accurate and supported by the retrieved evidence. These evaluations do not support analysis of LLM agents' ability to develop RAG systems. Second, automated RAG optimization methods typically operate within a predefined space of pipeline components and hyperparameter settings~\cite{kim2024autorag,fu2024autorag,orbach2025raghpo}, even when an LLM helps determine which configurations to evaluate~\cite{zhou2026hammer,strand2026agentic}. In this paper, we propose AutoRAGBench, to our knowledge the first benchmark for evaluating how effectively LLM agents can automate RAG system development in an open code space.

To construct AutoRAGBench, we outline two challenges.
\textbf{C1: How can we design diverse tasks and clean, isolated environments to fairly assess LLM agents’ ability to build RAG systems autonomously?}
Different corpora and question types call for different RAG strategies, so a comprehensive assessment of LLM agents’ ability to automate RAG system development requires tasks that span a range of settings.
For a fair comparison, each agent needs a clean, isolated environment with the same configuration and the freedom to write code, install libraries, and download models. Internet access also gives agents potential access to external datasets or leaked test answers, making it difficult to attribute performance gains to their design choices.
\textbf{C2: How can we evaluate not only the RAG system but also the agent's ability to refine it through iteration?}
The choice of answering model can affect answer accuracy, making it difficult to isolate the RAG system's contribution. The final score alone cannot show which improvements the agent made during development or whether those gains persisted across subsequent iterations.

To address C1, AutoRAGBench includes nine tasks across five categories.
The corpora cover domains such as news and medicine, and the questions include those that require multi-hop reasoning or fact verification.
Each build starts in a fresh container with its own workspace, while the test set stays outside the container.
Given the same task instructions, one H800 GPU, and six hours, agents can write code, install libraries, and download auxiliary models to build their systems from scratch.
Agents may use only the task data we provide and must not download external corpora or datasets.
We audit development trajectories and submissions for use of external data or leaked test answers.

To address C2, we evaluate both the RAG systems that agents deliver and how agents refine them during development.
For RAG system evaluation, five held-out answering models independently answer each test question using the same evidence retrieved by a given system.
We average answer scores across models, check whether system rankings change with the answering model, and measure retrieval latency.
For development process evaluation, we use development trajectories to examine how agents prepare for the task, implement retrieval methods, test their systems, and use test results to guide revisions.
Within each build, we compare the last observed development score with the first and the best comparable scores to measure overall progress and determine whether earlier gains persist.

\textbf{Our findings.}
GPT-6.1 Sol with Codex CLI achieves the highest average answer accuracy across nine tasks, 56.1 points.
It exceeds Reference RAG by 9.7 points and the composite Human-SOTA baseline under our protocol by 11.7 points.
It also improves seven existing human-designed systems by 16.6 points on average while preserving their core mechanisms.
In our adapted comparisons, it also outperforms automated searches over predefined RAG pipelines~\cite{strand2026agentic,jin2025flashrag}, including a search driven by the same LLM.
Together with the observed task-specific designs, these results suggest value in adapting retrieval and evidence-selection logic beyond a fixed set of configurations.
Across 77 builds with comparable development scores, 72 reach scores at least one point above their first version.
The median first-to-last gain is 5.5 points.
Yet 20 builds have last observed development scores at least one point below their best.
Larger development gains do not always correspond to higher hidden-test answer accuracy.

The contributions of this paper are summarized as follows:
\begin{itemize}
\item We propose AutoRAGBench, the first benchmark to evaluate how well LLM agents automate the development of RAG systems from scratch in an open code space.
\item We design an evaluation framework that combines RAG system evaluation with development process evaluation to assess how well the systems perform and how effectively agents improve them through iteration.
\item We evaluate nine agents across nine tasks and provide a detailed analysis of the experimental results.
We release our code and data on \href{https://github.com/ZhanGHanG9991/AutoRAGBench}{GitHub}\footnote{\url{https://github.com/ZhanGHanG9991/AutoRAGBench}} and our \href{https://zhanghang9991.github.io/AutoRAGBench_Website}{project website}\footnote{\url{https://zhanghang9991.github.io/AutoRAGBench_Website}} to support further research.
\end{itemize}
\section{Preliminaries}
\label{sec:preliminaries}

In this section, we review RAG fundamentals and formalize the RAG system development task in AutoRAGBench.

\subsection{RAG}
RAG enables an LLM to answer questions using knowledge retrieved from external knowledge sources~\cite{lewis2020retrieval}.
Given a question $q$, a retriever $R$ retrieves relevant knowledge fragments $\mathcal{Z}_q$ from an external knowledge source $\mathcal{C}$.
An LLM-based generator $G$ then produces an answer using the question and the retrieved knowledge:
\begin{equation}
    \mathcal{Z}_q = R(q;\mathcal{C}), \qquad \hat{y} = G(q,\mathcal{Z}_q).
\end{equation}

RAG system design involves choices such as retrieval strategies, context construction, and control flow.
These choices affect how the system retrieves and uses knowledge, as well as its answer quality and runtime efficiency.

\subsection{Task Formulation}
An AutoRAGBench task asks an LLM agent to build a RAG system from scratch using a given external knowledge source and question--answer examples under resource budgets and rules.
In an open code space, the agent iteratively designs the architecture, writes code, runs tests, and uses observations to refine the system.
We represent the task specification as
\begin{equation}
    \mathcal{T} = \left\langle
        \mathcal{C},\,
        \mathcal{D},\,
        \mathcal{B},\,
        \mathcal{K}
    \right\rangle,
\end{equation}
where $\mathcal{C}$ is the external knowledge source and $\mathcal{D}=\{(q_i,y_i)\}_{i=1}^{n}$ contains $n$ question--answer examples for $\mathcal{C}$, with question $q_i$ and reference answer $y_i$.
The LLM agent may use these examples to develop and validate the RAG system.

The resource budget $\mathcal{B}$ specifies the time, compute, memory, and storage available for system development.
The rule set $\mathcal{K}$ specifies rules for data access, network access, and model use, along with requirements for the submitted RAG system, such as its input and output interfaces.

We represent the LLM agent as
\begin{equation}
    a = \langle M_a,\mathcal{H}_a\rangle,
\end{equation}
where $M_a$ is the underlying LLM and $\mathcal{H}_a$ is the harness that supports the LLM agent's operation.
The LLM forms plans and selects actions from the current context; the harness assembles model inputs, coordinates tool calls, manages the execution loop, and returns observations to the LLM.

Let $\mathcal{A}_a$ denote the action space available to the LLM agent under the task rules.
It includes operations such as reading, writing, and editing files, searching file contents, and executing shell commands and scripts.
Through these actions, the LLM agent interacts with a development environment $\mathcal{E}$.
Let $h_0=\emptyset$ denote the initial interaction history.
At iteration $t$, the LLM agent selects an action based on the task specification and the preceding history:
\begin{equation}
    u_t = \pi_a(\mathcal{T},h_{t-1}),
    \qquad u_t\in\mathcal{A}_a,
\end{equation}
where $\pi_a$ denotes the action-selection process.
The harness invokes the corresponding tool to execute the action in $\mathcal{E}$ and collects the resulting observation $o_t$.
We express the observation and updated history as follows:
\begin{equation}
\begin{aligned}
    o_t &= \mathcal{E}(u_t,h_{t-1}),\\
    h_t &= h_{t-1}\oplus\langle u_t,o_t\rangle,
\end{aligned}
\end{equation}
where $\oplus$ denotes appending the action--observation pair to the history.
The history $h_t$ records interactions through iteration $t$.
The LLM agent iteratively develops the system based on these observations, subject to the resource budget $\mathcal{B}$ and rule set $\mathcal{K}$.
We denote the iterative development process by
\begin{equation}
    S = \operatorname{Build}(a;\mathcal{T}),
\end{equation}
where $S$ is the final RAG system, which implements the retrieval function $R$.
The agent aims to retrieve evidence that supports accurate answers with low retrieval latency:
\begin{equation}
    \underset{S}{\operatorname{maximize}}\; Q(S),
    \qquad
    \underset{S}{\operatorname{minimize}}\; L(S),
\end{equation}
where $Q(S)$ is the downstream answer score and $L(S)$ is the retrieval latency.

\section{AutoRAGBench}

In this section, we detail the construction of AutoRAGBench, including its task suite, system contract, build phase, and integrity audit.

\subsection{Overview}

\begin{figure*}[t]
    \centering
    \includegraphics[width=\textwidth]{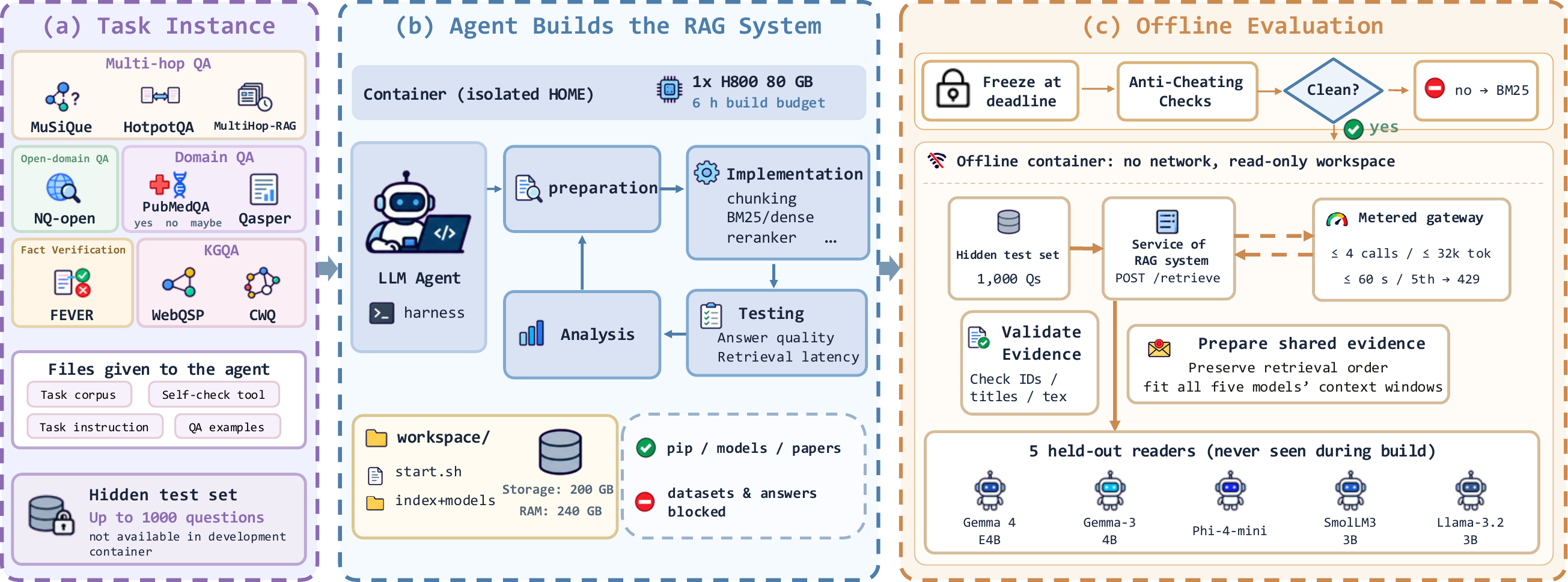}
    \caption{AutoRAGBench pipeline.}
    \label{fig:benchmark-overview}
\end{figure*}

Figure~\ref{fig:benchmark-overview} shows the pipeline of AutoRAGBench.
We assign each agent a target benchmark and provide the benchmark's external knowledge source and question--answer examples from its training split.
Each agent operates in a clean container with one H800 GPU and internet access during development.
The agent has six hours to build an accurate and efficient RAG system from scratch, without starter code and prescribed designs.
Within the resource budget and rule set, agents autonomously design, implement, validate, and iteratively refine their RAG systems.
The rule set restricts agents to the supplied external knowledge source for retrieval and the supplied question--answer examples for development and validation. This ensures consistent data availability across agents and enables a fair comparison of the RAG systems they build.

For evaluation, we first use LLM-based audits to check whether agents follow the rule set throughout development.
We evaluate the RAG system on the target benchmark's hidden test set only if the run passes the audit.
For each test question, five answering models independently generate answers from the same evidence retrieved by the system.
We report the downstream answer score and retrieval latency of the system.
We compute the overall score based on results from five answering models and nine benchmarks.

\subsection{Task Suite}
\label{sec:task-suite}

As shown in Table~\ref{tab:task-suite}, the task suite in AutoRAGBench consists of nine benchmarks spanning open-domain question answering (QA), document-based multi-hop QA, domain-specific QA, fact verification, and knowledge-graph QA.
For each benchmark, we retain up to 10,000 training examples and 1,000 test examples to ensure comparable evaluation scales across tasks.
We report corpus sizes as record counts, with record types varying across benchmarks (e.g., passages, full papers, and news articles).

\begin{table}[t]
\centering
\caption{Task suite statistics and evaluation metrics.}
\label{tab:task-suite}
\begingroup
\footnotesize
\setlength{\tabcolsep}{2.5pt}
\renewcommand{\arraystretch}{1.12}
\begin{tabular*}{\columnwidth}{@{\extracolsep{\fill}}l l r r r l@{}}
\toprule
\textbf{Category} & \textbf{Benchmark} & \textbf{\#Records} & \textbf{Train} & \textbf{Test} & \textbf{Metric} \\
\midrule
Open-domain QA & NQ & 21,015,324 & 10,000 & 1,000 & EM \\
\midrule
\multirow{3}{*}{\shortstack[l]{Document-based\\multi-hop QA}} & HotpotQA & 5,233,235 & 10,000 & 1,000 & F1 \\
 & MuSiQue & 117,534 & 10,000 & 1,000 & F1 \\
 & MH-RAG & 609 & 1,680 & 500 & EM \\
\midrule
Domain-specific QA & PubMedQA & 273,518 & 500 & 500 & Acc. \\
 & QASPER & 1,585 & 3,598 & 1,000 & F1 \\
\midrule
Fact verification & FEVER & 5,396,106 & 10,000 & 1,000 & Acc. \\
\midrule
Knowledge-graph QA & WebQSP & 665,334 & 2,945 & 1,000 & Hits@1 \\
 & CWQ & 1,310,816 & 10,000 & 1,000 & Hits@1 \\
\bottomrule
\end{tabular*}
\endgroup
\end{table}

\noindent\textbf{Open-domain QA.}
NQ-open (NQ)~\cite{Lee2019LatentRF}, the open-domain version of Natural Questions~\cite{Kwiatkowski2019NaturalQA}, pairs real Google search queries with short factual answers.
We use Wikipedia passages~\cite{Karpukhin2020DensePR} as the retrieval corpus for finding supporting evidence.

\noindent\textbf{Document-based multi-hop QA.}
We select three benchmarks that require multi-hop reasoning over document collections.
HotpotQA~\cite{Yang2018HotpotQAAD} features bridge and comparison questions that require reasoning across two Wikipedia articles.
We use its hard questions and its fullwiki corpus, which consists of the introductory paragraphs of 5.2M Wikipedia articles.
MuSiQue~\cite{Trivedi2021MM} composes questions from two to four connected single-hop questions.
We use MuSiQue-Ans, its answerable-only version, and pool the released paragraphs from all splits into a deduplicated retrieval corpus.
MultiHop-RAG (MH-RAG)~\cite{Tang2024MultiHopRAGBR} tests inference, comparison, and temporal reasoning across news articles.
We retain article-level documents with metadata and require abstention on unanswerable questions.

\noindent\textbf{Domain-specific QA.}
PubMedQA~\cite{Jin2019PubMedQAAD} contains biomedical research questions answered with \emph{yes}, \emph{no}, or \emph{maybe}.
We use its expert-labeled questions and build the corpus from abstracts across PubMedQA subsets.
QASPER~\cite{Dasigi2021ADO} contains questions about NLP papers, including questions that the target paper cannot answer.
We retain each paper's full text as a single document, and each question specifies its target paper, following the original setting.

\noindent\textbf{Fact verification.}
FEVER~\cite{Thorne2018FEVERAL} requires classifying claims as \emph{Supports}, \emph{Refutes}, or \emph{NotEnoughInfo} based on Wikipedia evidence.
We retrieve from the introductory sections of the 5.4M pages in the official FEVER Wikipedia dump.

\noindent\textbf{Knowledge-graph QA.}
WebQSP~\cite{Yih2016TheVO} contains questions answerable within one or two Freebase relations of a topic entity.
ComplexWebQuestions (CWQ)~\cite{Talmor2018TheWA} extends WebQSP queries with compositions, conjunctions, comparatives, and superlatives, requiring up to four hops.
For each benchmark, we merge the Freebase subgraphs released by RoG~\cite{Luo2023ReasoningOG} into a single knowledge graph, provided both as triples and as one document per entity listing its outgoing relations and targets.
Each question is given with the names of its topic entities.

We use each benchmark's standard evaluation metric.
NQ and MH-RAG use exact match (EM), which requires the prediction to match any reference answer exactly.
HotpotQA, MuSiQue, and QASPER use answer token F1 to measure token overlap between the prediction and reference answers.
FEVER and PubMedQA use label accuracy (Acc.).
WebQSP and CWQ use Hits@1, which checks whether the single predicted entity or value matches any reference answer.
For EM, F1, and Hits@1, we lowercase answers, remove punctuation and English articles, and normalize whitespace.
When a question has multiple reference answers, we report the maximum F1 score across all reference answers.

\subsection{System Development Protocol}
\label{sec:build-phase}

The system development protocol specifies the development setup, the requirements that submitted systems must satisfy, and how the evaluator freezes submissions before evaluation.

Each build runs in a clean and isolated container, with no persistent agent memory carried over between runs. Each agent receives the benchmark's corpus and question-answer examples from its training split, which it may divide into subsets for system development and validation. The test set remains outside the container and unavailable to the agent during development. All agents receive the same task instructions, one H800 GPU, and six hours to develop their systems. If an agent finishes early with sufficient time remaining, we prompt it to continue development. All agents receive the same prompt.

Each system must use only the supplied corpus for indexing and evidence retrieval. Agents build RAG systems from scratch, without starter code, a prescribed pipeline, or a predefined configuration space. They autonomously explore and refine designs for tasks such as corpus processing, indexing, retrieval, and reranking, while also choosing prompts and control flow. Systems may use auxiliary models within the hardware limits. Given a question \(q\), each system must return an ordered list of evidence spans from the supplied corpus.
Generated summaries, hypothetical documents, and intermediate answers may guide retrieval but cannot replace corpus evidence in the returned list.
The evaluator validates document identifiers, titles, and text against the supplied corpus before scoring.

Before evaluation, we freeze the submitted RAG system. During evaluation, we prohibit further training and implementation changes, and require each test question to be processed independently of the others.

\subsection{Anti-Cheating Checks}
\label{sec:anti-cheating-checks}

As an agent develops its RAG system, we record its actions and observations at each step. We call this sequence of records the run's development trajectory.

We audit development trajectories and submitted RAG systems for the use of additional corpora, downloads of benchmark test sets, and other rule violations. We use DeepSeek to automatically check these trajectories and systems for potential violations. Our team then manually examines runs whose trajectories suggest potential violations.
When we confirm a violation, we rerun the experiment to give the agent another chance to develop a compliant RAG system. If the new system passes the audit, we evaluate it; otherwise, we assign the run the naive BM25 RAG baseline score.
\section{Evaluation Design}
\label{sec:evaluation-design}

In this section, we present AutoRAGBench's evaluation design, including RAG system evaluation and development process evaluation.
AutoRAGBench evaluates both the RAG systems that agents deliver and the development processes that produce the systems.
On one hand, RAG system evaluation assesses the performance and runtime efficiency of the systems agents build across tasks.
Final performance alone does not show how much the system improves during development or whether the agent retains earlier gains.
On the other hand, development process evaluation examines how effectively agents improve the RAG systems through iteration.

\subsection{RAG System Evaluation}
\label{sec:outcome-evaluation}

We use answer accuracy to assess how well the evidence each RAG system retrieves supports correct answers.
We examine rank stability to see whether the choice of answering model affects how systems compare.
We measure retrieval latency from the time a system receives a question until it returns the retrieved evidence.

\noindent\textbf{Answer accuracy.}
We evaluate each final RAG system by scoring answers generated from its retrieved evidence.
For agent $a$ and task $\mathcal{T}_k$, $S_{a,k}=\operatorname{Build}(a;\mathcal{T}_k)$ denotes the final RAG system.
Its retrieval function $R_{a,k}$ retrieves evidence from the task's knowledge source $\mathcal{C}_k$.
Let $\mathcal{D}^{\mathrm{test}}_k=\{(q_j,y_j)\}_{j=1}^{n_k}$ denote the hidden test set, where $q_j$ is a question and $y_j$ is its reference answer.
For each question $q_j$, let $\mathcal{Z}_{q_j}$ denote the evidence returned by $R_{a,k}$ and $\hat{y}_j$ the answer that model $G$ generates from this evidence.
\begin{equation}
\begin{aligned}
    \mathcal{Z}_{q_j}
        &=R_{a,k}(q_j),\\
    \hat{y}_j
        &=G\left(q_j,\mathcal{Z}_{q_j}\right).
\end{aligned}
\end{equation}
The five held-out answering models $G\in\mathcal{G}$ are used only for final evaluation; development checks use Qwen3.5-4B.
Before any held-out model answers, the evaluator validates and packs the retrieved evidence into one common prefix that fits every model's context window (Section~\ref{sec:evaluation-protocol}).
For each system we compare, the same model answers the test questions using the evidence that system retrieves.
Then, we score $G$'s test answers against their reference answer using the task metric $\mu_k$, and average the scores to obtain $Q(S_{a,k};G)$.
\begin{equation}
    Q(S_{a,k};G)
    =\frac{1}{n_k}\sum_{j=1}^{n_k}\mu_k\left(\hat{y}_j,y_j\right).
\end{equation}

Let $\mathcal{G}$ denote a set of answering models, each generating answers from the same evidence retrieved by a given system.
We average the answer score $Q(S_{a,k};G)$ over the answering models in $\mathcal{G}$:
\begin{equation}
    \overline{Q}(S_{a,k};\mathcal{G})
    =\frac{1}{|\mathcal{G}|}\sum_{G\in\mathcal{G}}Q(S_{a,k};G).
\end{equation}

We report scores on a 0--100 scale.
The overall score is the unweighted mean of the nine task scores, so a task with more test questions does not receive greater weight.
This average combines each task's prescribed metric rather than treating all tasks as one pooled accuracy measure.

\noindent\textbf{Rank stability.}
We examine whether changing the answering model changes the relative ranking of RAG systems.
For each answering model $G$, we define the score-induced ordering over systems on the same task as
\begin{equation}
    S_{a,k}\succ_G S_{b,k}
    \iff Q(S_{a,k};G)>Q(S_{b,k};G).
\end{equation}
Systems with equal scores are tied.
We assess rank stability by comparing these orderings across answering models.

\noindent\textbf{Retrieval latency.}
Retrieval latency $\ell(S_{a,k};q_j)$ measures the time from when system $S_{a,k}$ receives question $q_j$ until it returns the retrieved evidence.
This latency excludes final answer generation by $G$.
We define the mean retrieval latency over the test set as
\begin{equation}
    L(S_{a,k})
    =\frac{1}{n_k}\sum_{j=1}^{n_k}\ell(S_{a,k};q_j).
\end{equation}
We compare systems on both $\overline{Q}(S_{a,k};\mathcal{G})$ and $L(S_{a,k})$ to assess the trade-off between answer accuracy and retrieval latency.

\subsection{Development Process Evaluation}
\label{sec:process-evaluation}

We examine four development activities to understand how agents build their systems and act on feedback.
We assess progress across iterations to see whether agents improve their systems and retain earlier gains.
For each activity, we measure its share of total development time and tokens. We also track how API spending varies across tasks.

\begin{figure*}[t]
    \centering
    \includegraphics[width=\textwidth]{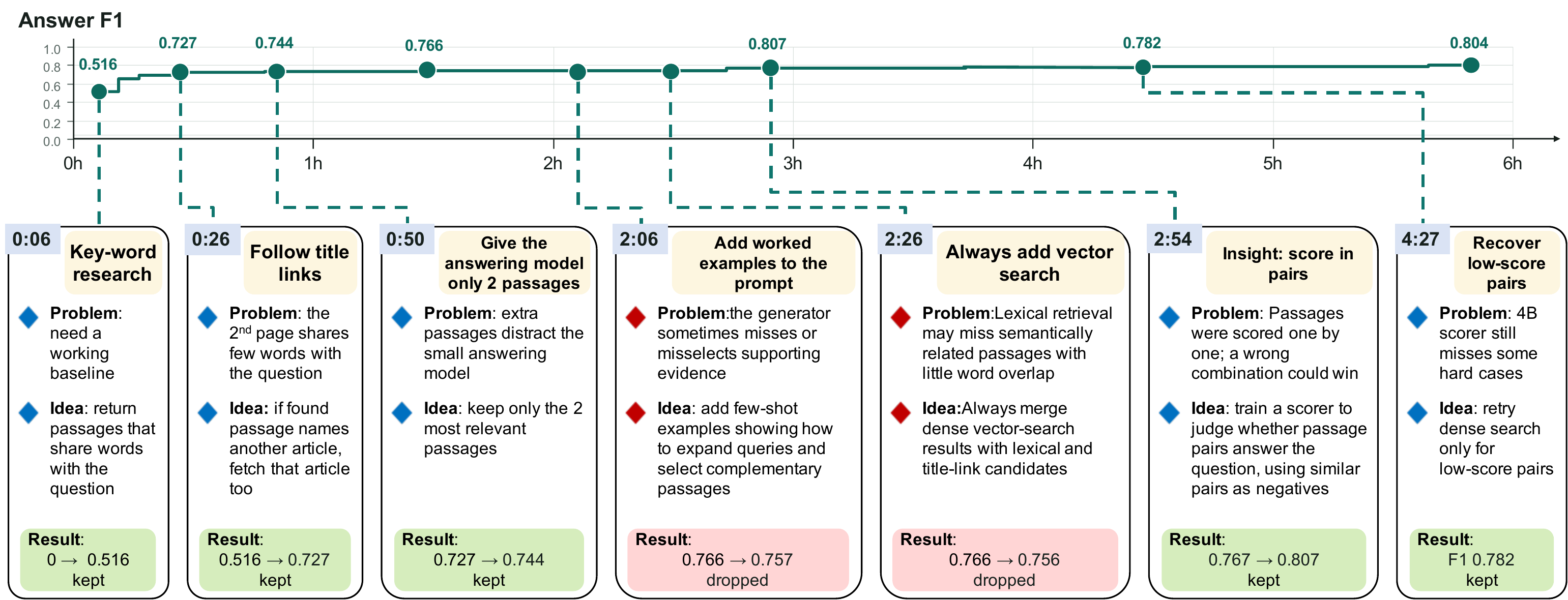}
    \caption{\textbf{A HotpotQA development trace.} GPT-6.1 Sol (Codex CLI) builds RAG system within a six-hour budget.}
    \label{fig:build-case}
\end{figure*}

\noindent\textbf{Development activities.}
We use development trajectories and submitted systems to characterize four activities.

\noindent\ding{182}~\textit{Preparation.}
We identify when an agent examines the task inputs and its working environment.
These checks help the agent determine how the task data are organized and which tools it can use to build the retriever.

\noindent\ding{183}~\textit{Implementation.}
We examine which RAG methods agents implement for each task, such as query rewriting, graph traversal, reranking, and evidence selection.
We assess whether agents choose methods suited to the task and build the components those methods require, for example by constructing sparse or dense indexes or training rerankers.

\noindent\ding{184}~\textit{Testing.}
We examine whether agents test the RAG systems they build or individual components of those systems, which questions they select, and whether they use separate development questions for tuning and validation.

\noindent\ding{185}~\textit{Analysis.}
We examine how agents interpret test results to diagnose problems in their RAG systems and decide what to try next.
This helps us assess whether agents can propose changes that address the specific problems they identify.

\noindent\textbf{Progress across iterations.}
The goal is to assess whether agents improve their RAG systems through iteration and whether they retain earlier gains.
Figure~\ref{fig:build-case} follows the agent across iterations as it tests changes to the HotpotQA retriever and decides which to keep.
For a single build, let $d_1,\ldots,d_N$ denote comparable development scores for $N\geq 2$ observed versions in chronological order.
We measure the change in score from the first observed version to the last ($\Delta_{\mathrm{dev}}$) and the drop from the highest observed score to the last ($\Delta_{\mathrm{drop}}$).
\begin{equation}
\begin{aligned}
    \Delta_{\mathrm{dev}} &= d_N-d_1,\\
    \Delta_{\mathrm{drop}} &= \max_{1\leq i\leq N}d_i-d_N.
\end{aligned}
\end{equation}
A zero $\Delta_{\mathrm{drop}}$ means that the last observed score matches the best observed score.
These are within-build descriptive measures: question sets can differ across builds, and the last observed check need not evaluate the exact system frozen at the deadline.
We examine these two measures alongside the score trajectories and the frequency of iterations in which the agent modifies the system and then evaluates it.

\noindent\textbf{Development resource use.}
We measure how much time each development activity takes and how much each agent spends on API calls across tasks.
We use output token counts to quantify the volume of model output for each activity.

\section{Experimental Setup}

In this section, we detail the evaluated LLM agents, baselines, measurements, and execution settings for experiments.

\subsection{Evaluated LLM Agents}
\label{sec:evaluated-agents}

We evaluate nine LLM agents in the experiments.
Whenever available, we use the native agent harness developed by each LLM provider, thereby evaluating each model under its provider-recommended setup.
When a native harness is unavailable, we use a harness commonly paired with the model.
We use four OpenAI models (GPT-6.1 Sol~\cite{openai2026gpt61sol}, GPT-6 Sol~\cite{openai2026gpt6sol}, GPT-5.6 Sol~\cite{openai2026gpt56sol}, and GPT-5.6 Luna~\cite{openai2026gpt56luna}) with Codex CLI~\cite{openai2026codex}.
We use DeepSeek V4.1 Flash\footnote{We use DeepSeek-V4.1-Flash-1017~\cite{deepseek2026v41flash} in the experiments.} with DeepSeek Harness~\cite{deepseek2026harness}.
For GLM-5.3~\cite{zai2026glm53}, GLM-5.3-Flash~\cite{zai2026glm53flash}, Qwen3.8 Flash~\cite{qwen2026qwen38flash}, and Gemini 3.8 Flash~\cite{google2026gemini38flash}, whose providers do not offer dedicated harnesses, we use Claude Code~\cite{anthropic2026claudecode} for the first three models and OpenCode~\cite{anomaly2026opencode} for Gemini 3.8 Flash~\cite{terminalbench2026leaderboard4,pmlr-v306-rank26a}.

\subsection{Answering Models}

An answering model takes a question and the evidence retrieved by the RAG system as input and generates an answer.
We use five language models as answering models: Gemma-4-E4B~\cite{Abd2026Gemma4T}, Gemma-3-4B~\cite{Kamath2025Gemma3T}, Phi-4-mini~\cite{Abouelenin2025Phi4MiniTR}, SmolLM3-3B~\cite{bakouch2025smollm3}, and Llama-3.2-3B~\cite{llama32}.
We select models from four families with different parameter counts to assess the robustness of system comparisons across answering models.

\subsection{Baselines}

\noindent\textbf{Closed-book.}
The answering models answer each question without evidence, relying only on their parametric knowledge.

\noindent\textbf{Naive BM25 RAG.}
This baseline splits the corpus into passages of about 200 words, indexes them with BM25~\cite{Robertson1994OkapiAT}, and returns the top five passages for each question.
We assign this baseline's score to any run that still fails the anti-cheating checks after a rerun.

\noindent\textbf{Reference RAG.}
This baseline combines BM25 and bge-base-en-v1.5~\cite{bge_embedding} dense retrieval over the same passages using reciprocal rank fusion, reranks the candidates with bge-reranker-base, and returns the top eight passages.
On multi-hop tasks, it uses Qwen3.5-4B~\cite{qwen3.5} as an auxiliary LLM to generate one follow-up query and retrieve additional evidence.

\noindent\textbf{Human-SOTA.}
We include a set of recent, representative human-designed RAG methods.
These methods cover several technical directions, including graph-based indexing (HippoRAG 2~\cite{pmlr-v267-gutierrez25a}, LightRAG~\cite{Guo2024LightRAGSA}, LinearRAG~\cite{Zhuang2025LinearRAGLG}, LEGO-GraphRAG~\cite{Cao2024LEGOGraphRAGMG}, and Clue-RAG~\cite{Su2026ClueRAGTA}), knowledge-graph-based subgraph retrieval (SubgraphRAG~\cite{li2024subgraphrag}), hierarchical indexing (RAPTOR~\cite{sarthi2024raptor} and BookRAG~\cite{10.14778/3819518.3819556}), query rewriting and retrieval control (HyDE~\cite{gao-etal-2023-precise}, CRAG~\cite{yan-etal-2026-corrective}, and Adaptive-RAG~\cite{Jeong2024AdaptiveRAGLT}), and automatic tuning of RAG configurations (HAMMER~\cite{zhou2026hammer} and CARROT~\cite{11629401}).
To ensure a fair comparison, we adapt each method based on its official implementation and require it to use only the designated generation model at test time and comply with the benchmark's input/output interface specification.
We impose a 24-hour time limit on system development for each task. Any method that exceeds this limit is assigned a score of zero for the corresponding task.
To ensure compatibility with our evaluation protocol, we make the necessary adaptations, including constructing graphs from the task corpus, disabling web search, and capping the number of iterative retrieval calls at the specified limit.
We further construct a composite Human-SOTA baseline by selecting the highest score achieved by any of these methods on each task.
These are adapted implementations evaluated under our corpus, service, and answering-model constraints.
Their 24-hour development limit differs from the agents' six-hour budget.
The composite is a per-task envelope over these implementations, not a single deployable system or an upper bound on human-designed RAG.
Its scores need not reproduce the results reported in the original papers.

\noindent\textbf{Automated RAG baselines.}
We compare agents with Agentic AutoRAG~\cite{strand2026agentic} and FlashRAG~\cite{jin2025flashrag}.
Agentic AutoRAG uses a development LLM to iteratively configure chunking, embedding, retrieval, query expansion, and reranking.
Its deployed service uses only the fixed generation model.
The FlashRAG baseline automatically evaluates predefined single-pass and multi-step retrieval pipelines and selects the one with the best validation performance.

\subsection{Cost Analysis}
\label{sec:api-cost}
We report total API costs for agents building RAG systems across nine tasks.
GPT-5.6 Sol has the highest total at $\sim$\$487; GPT-6 Sol costs $\sim$\$450.
Gemini 3.8 Flash, GPT-6.1 Sol, and GLM-5.3 cost $\sim$\$195, \$167, and \$107, respectively.
GPT-5.6 Luna and GLM-5.3-Flash cost $\sim$\$21 and \$20.
DeepSeek V4.1 Flash costs $\sim$\$9.5, and Qwen3.8 Flash costs $\sim$\$6.

\subsection{Implementation Details}
\label{sec:evaluation-protocol}

\noindent\textbf{Execution environment.}
In the main experiment, each agent runs in a clean, isolated Debian 12 container with one NVIDIA H800 GPU, 12 CPU cores, 240\,GB of host memory, and a 200\,GB workspace limit.

\noindent\textbf{System development and evaluation.}
Agents build from scratch and use \texttt{self\_check} on subsets of the supplied training examples.
Qwen3.5-4B~\cite{qwen3.5} answers development questions and supports generation during retrieval.
We evaluate the system frozen at six hours. Agents must restore any preferred earlier version by that deadline.

\noindent\textbf{Retrieval interface and budgets.}
The \texttt{POST /retrieve} endpoint returns ordered corpus passages with document identifiers, titles, and original text.
Per question, the gateway allows at most four Qwen3.5-4B calls and 32,000 total tokens within 60 seconds.
The service cannot access the five held-out answering models.
The evaluator verifies passages and packs them in order into a common prefix of at most 32,000 characters and 12,000 Qwen tokens, truncating the last passage if needed.
The prefix must fit all five models' 32,768-token contexts with 512 tokens reserved for each answer.
Invalid evidence, budget violations, and failed generations count as invalid predictions.

\noindent\textbf{Answering model settings.}
We use zero-shot prompting for all benchmarks and apply each answering model's native chat template. We use the same task-specific answer instruction across models, set the temperature to 0, and limit output to 512 tokens.

\begin{table*}[t]
\centering
\caption{\textbf{Main results}, ranked by average score.
Scores are primary metrics $\times$100 averaged over five held-out answering models; Avg is the mean over the nine tasks, and bold marks the best score on each task. DSH: DeepSeek Harness.}
\label{tab:main-results}
\begingroup
\footnotesize
\setlength{\tabcolsep}{0.5pt}
\renewcommand{\arraystretch}{1.2}
\begin{tabular*}{\textwidth}{@{\extracolsep{\fill}}c@{\hspace{3pt}}l*{10}{>{\centering\arraybackslash}p{36pt}}@{}}
\toprule
Rank & Method & \textbf{Avg} & NQ & HotpotQA & MuSiQue & MH-RAG & {\scriptsize PubMedQA} & QASPER & FEVER & WebQSP & CWQ \\
\midrule
1 & GPT-6.1 Sol (Codex CLI) & \textbf{56.08} & 38.60 & \textbf{65.47} & \textbf{32.91} & \textbf{63.36} & \textbf{63.48} & \textbf{48.27} & \textbf{58.82} & \textbf{78.92} & 54.92 \\
2 & GPT-6 Sol (Codex CLI) & 54.10 & 38.86 & 57.68 & 27.99 & 60.56 & 59.60 & 46.87 & 58.62 & 74.78 & \textbf{61.94} \\
3 & GPT-5.6 Sol (Codex CLI) & 51.21 & 38.62 & 52.41 & 26.87 & 54.44 & 62.48 & 46.03 & 57.24 & 75.86 & 46.98 \\
4 & GLM-5.3 (Claude Code) & 50.50 & \textbf{39.20} & 50.95 & 25.05 & 53.48 & 61.40 & 46.27 & 56.54 & 73.44 & 48.14 \\
5 & GLM-5.3-Flash (Claude Code) & 49.71 & 38.32 & 55.17 & 22.34 & 56.60 & 62.20 & 46.15 & 56.40 & 74.20 & 35.98 \\
6 & DeepSeek V4.1 Flash (DSH) & 49.41 & 38.90 & 51.65 & 21.48 & 58.36 & 62.76 & 45.65 & 55.78 & 70.42 & 39.70 \\
7 & Gemini 3.8 Flash (OpenCode) & 48.32 & 32.98 & 47.16 & 27.47 & 53.72 & 55.80 & 45.27 & 56.90 & 70.10 & 45.50 \\
8 & Reference RAG & 46.42 & 37.74 & 52.46 & 23.93 & 56.48 & 54.32 & 43.58 & 57.36 & 57.30 & 34.60 \\
9 & Qwen3.8 Flash (Claude Code) & 45.95 & 36.50 & 49.54 & 19.11 & 57.20 & 58.84 & 46.38 & 56.50 & 63.80 & 25.66 \\
10 & GPT-5.6 Luna (Codex CLI) & 45.61 & 32.96 & 47.12 & 22.31 & 53.60 & 58.00 & 45.92 & 56.66 & 61.12 & 32.76 \\
11 & Human-SOTA (composite) & 44.43 & 34.18 & 41.05 & 22.64 & 52.24 & 57.64 & 45.29 & 58.08 & 54.18 & 34.58 \\
12 & Naive BM25 RAG & 36.66 & 24.86 & 35.79 & 15.37 & 54.48 & 51.96 & 40.61 & 55.90 & 30.64 & 20.38 \\
13 & Closed-book & 25.30 & 13.36 & 21.98 & 8.55 & 38.92 & 38.20 & 18.46 & 44.40 & 24.82 & 19.04 \\
\bottomrule
\end{tabular*}
\endgroup
\end{table*}

\noindent\textbf{Runtime cost measurement.}
We cap retrieval latency at 60 seconds per question and record 60 seconds for timeouts and missing measurements.
We compute the mean latency for each task and then average the nine task means.
During runtime, the gateway records the time spent on each operation and the agent's API spending.

\subsection{Environment}
\label{sec:environment}

The main experiment uses two Ubuntu 22.04.4 LTS nodes with 16 NVIDIA H800 (80\,GB) GPUs in total.
Each node has 104 CPU cores across two Intel Xeon Platinum 8470 processors and 1 TB of host memory.

\section{Experimental Results}

Following the evaluation design in Section~\ref{sec:evaluation-design}, we evaluate both the RAG systems that agents build and the processes they use to develop them.
We then examine whether agents can improve human-designed methods and how they compare with automated RAG frameworks.

\subsection{Main Results}

Table~\ref{tab:main-results} reports answer accuracy averaged over five held-out answering models.
Across the nine tasks, closed-book answering averages 25.3 points, and naive BM25 RAG reaches 36.7.
The reference RAG system combines hybrid retrieval with reranking and reaches 46.4.
GPT-6.1 Sol with Codex CLI achieves the highest average, 56.1, and the best score on seven tasks.
Its 9.7-point gain over the reference system is comparable to the gain from naive BM25 RAG to the reference system.
All nine agent configurations outperform naive BM25 RAG, and seven outperform the reference system.
The composite Human-SOTA baseline averages 44.4, below all nine agent configurations.
GPT-6.1 Sol exceeds this baseline by 11.7 points.

On WebQSP, all nine configurations outperform the reference system, by 14.1 points on average.
Seven also outperform it on CWQ, with a maximum gain of 27.3 points.
All configurations improve on the reference system on the two domain-specific tasks, PubMedQA and QASPER.
Their average gains are 6.2 and 2.7 points, respectively.
On HotpotQA, the best configuration gains 13.0 points, but only three outperform the reference system.
The mean gain is just 0.6 points.
On NQ and FEVER, the largest gain is only 1.5 points on each task.
The knowledge-graph and domain-specific tasks therefore show gains across most configurations.
HotpotQA combines a large gain for the best configuration with a small mean gain across agents.

Across agents, the score range is 36.3 points on CWQ, 18.4 on HotpotQA, and 17.8 on WebQSP.
On QASPER and FEVER, it is only about 3.0 points each.
QASPER therefore shows broad gains over the reference system but little separation among agents.
CWQ and HotpotQA reveal larger differences between the systems that agents build.
Only GPT-6.1 Sol and GPT-6 Sol outperform the reference system on all nine tasks.
On MH-RAG, four agent-built systems score below naive BM25 RAG.
A broad task suite captures these distinct aspects: overall gains, differences between agents, and weaknesses on individual tasks.

With Codex CLI, GPT-6.1 Sol scores 2.0 points higher than GPT-6 Sol on average.
The three document-based multi-hop tasks account for 1.7 points of this nine-task average gap.
Yet GPT-6 Sol scores 7.0 points higher on CWQ.
Knowledge-graph tasks drive broad gains over the reference system.
Document-based multi-hop tasks account for most of the gap between the two leading configurations.
GPT-5.6 Sol also outperforms GPT-5.6 Luna by 5.6 points on average under the same harness.
These comparisons show differences in both overall performance and task-specific strengths between LLMs under a fixed harness.
Cross-harness rankings compare complete agent configurations because both the LLM and harness change.

\subsection{Stability Across Answering Models}
\label{sec:reader-stability}

\begin{figure}[t]
    \centering
    \includegraphics[width=\columnwidth]{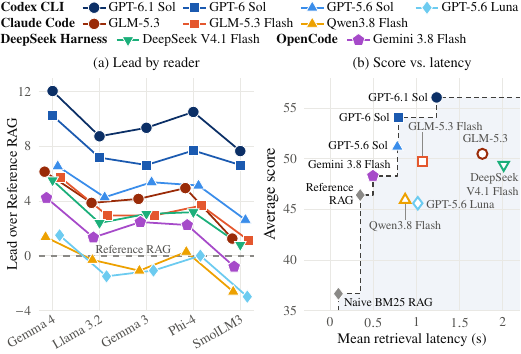}
    \caption{\textbf{Retrieval performance across answering models and the score–latency trade-off.} (a) Nine-task average score gain over reference RAG by answering model. (b) Average score vs. mean retrieval latency, with the dashed line marking the Pareto frontier.}
    \label{fig:reader-latency}
\end{figure}

Figure~\ref{fig:reader-latency}(a) plots each configuration's nine-task average under one answering model, minus the reference system's average under the same answering model.
GPT-6.1 Sol ranks first under all five answering models, and the top four configurations keep identical ranks.
The lines stay nearly parallel: the gap between the best and worst configurations remains 10.2 to 10.7 points.
The lead over the reference system changes more, with a mean of 1.5 to 5.9 points and six to nine configurations above it.
Rankings on most individual tasks vary more than this average.

\subsection{Efficiency}

Figure~\ref{fig:reader-latency}(b) plots mean answer accuracy against mean retrieval latency across the nine tasks.
Every agent-built system takes longer on average to retrieve evidence than Reference RAG.
Mean retrieval latency ranges from 0.50 to 2.01 seconds per question.
This is 1.4 to 5.8 times the reference system's mean of 0.35 seconds.
Naive BM25 RAG takes 0.10 seconds per question.
We measure retrieval latency from receiving the question to returning the evidence.
It excludes final answer generation by the answering model.
Query rewriting, multi-step retrieval, and reranking all count toward retrieval latency.
This includes the model calls these steps require.
On HotpotQA, for example, the GPT-6.1 Sol retriever uses title-bridge expansion and a learned passage-pair scorer.
It falls back on dense retrieval when confidence is low (Figure~\ref{fig:build-case}).

Naive BM25 RAG, Reference RAG, and Gemini 3.8 Flash with OpenCode lie on the Pareto frontier.
The three Sol configurations with Codex CLI also lie on it.
Along the frontier, the latency cost of each additional point varies.
Gemini 3.8 Flash gains 1.9 points over Reference RAG for about 0.15 additional seconds per question.
GPT-5.6 Sol adds another 2.9 points for about 0.28 additional seconds.
GPT-6 Sol then gains a further 2.9 points at nearly unchanged latency, from 0.78 to 0.79 seconds.
GPT-6.1 Sol gains another 2.0 points as latency rises from 0.79 to 1.23 seconds.
This is the highest latency cost per point on the frontier.

For each of the other five configurations, at least one system retrieves evidence faster and scores higher.
GLM-5.3-Flash takes 1.07 seconds per question, while GPT-5.6 Sol and GPT-6 Sol both retrieve evidence faster and score higher.
GLM-5.3 takes 1.76 seconds per question and trails all three Sol configurations in both retrieval speed and average score.
DeepSeek V4.1 Flash takes the longest at 2.01 seconds per question and scores below all three Sol and both GLM configurations.
GPT-5.6 Luna takes 2.9 times as long as Reference RAG and scores 0.8 points lower.
Qwen3.8 Flash takes 2.5 times as long as Reference RAG and scores 0.5 points lower.
Gemini 3.8 Flash also retrieves evidence faster than Luna and Qwen and scores 2.4 to 2.7 points higher.
Extra retrieval time alone offers no guarantee that the evidence will support a better answer.
Retrieval latency also distinguishes systems with similar answer accuracy.
GLM-5.3-Flash scores just 0.3 points above DeepSeek V4.1 Flash, yet DeepSeek takes nearly twice as long per question.

\subsection{Strategies in the Final Systems}
\label{sec:final-strategies}

\begin{figure}[t]
    \centering
    \includegraphics[width=\columnwidth]{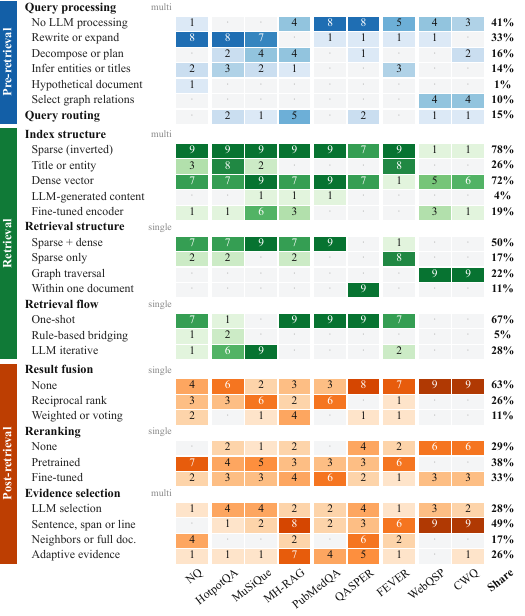}
    \caption{\textbf{Strategy use by task.} Cells count systems out of nine per task. Shares use all 81 systems except for retrieval flow, which excludes the 18 graph-traversal systems. Multi-choice groups can exceed 100\%.}
    \label{fig:strategy-by-task}
\end{figure}

\noindent\textbf{Coding the systems.}
Figure~\ref{fig:strategy-by-task} organizes the strategies of the 81 final systems by pipeline stage, following RAG taxonomies~\cite{huang2026survey,gao2023retrieval}.
We identify strategies from code, configurations, prompts, and agent reports, and check model use against gateway logs.
The task prompt suggests several strategies. A second coder has yet to verify the labels.

\noindent\textbf{Common foundation.}
Most systems combine classic indexes with a reranker: 63 use a sparse index, 58 use dense vectors, and 58 use a reranker, of which 27 are fine-tuned.
LLM use at indexing time is rare: only three systems index LLM-generated content, and only one uses a hypothetical document.
Another 33 systems perform no LLM query processing.

\noindent\textbf{Adaptation to tasks.}
Designs change markedly with the task.
All 18 knowledge-graph systems traverse the graph and return selected lines or paths, and 12 of them use no reranker; all nine QASPER systems rank within the identified paper; and eight FEVER systems use no dense vectors.
Query rewriting concentrates on NQ, HotpotQA, and MuSiQue, with eight, eight, and seven systems.
Iterative LLM retrieval concentrates on MuSiQue and HotpotQA, with nine and six systems, whereas all nine MH-RAG systems retrieve only once despite the task's multi-hop questions.
On MH-RAG, eight systems extract spans and seven adjust the evidence per question.

\noindent\textbf{Differences across configurations.}
The two leading configurations rely more on learned components and evidence control.
Across the nine tasks, GPT-6.1 Sol and GPT-6 Sol use learned classifiers or scorers on six and seven tasks, whereas no other configuration does so on more than three.
They adjust the evidence per question on seven and four tasks, again against at most three for any other configuration.
Fine-tuned rerankers do not follow this order: GLM-5.3-Flash uses them on five tasks, as many as GPT-6.1 Sol, whereas fourth-ranked GLM-5.3 uses them on one.
These counts describe strategy choices in the observed builds. Measuring their effects would require component ablations.

\subsection{Feedback and Iteration}
\label{sec:feedback-iteration}

\begin{figure}[t]
    \centering
    \includegraphics[width=\columnwidth]{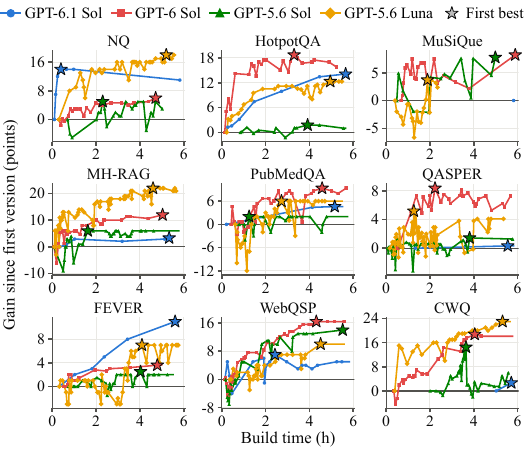}
    \caption{\textbf{Iteration curves for four Codex CLI configurations.} Each curve shows the change in Qwen3.5-4B score from the first evaluated version on a fixed set of questions for that build. Axis scales vary by task. Stars mark the earliest maximum.}
    \label{fig:iteration-gain}
\end{figure}

Across the nine configurations, 77 builds had at least two observed versions.
Their median net gain in development score from the first to the last observed version ($\Delta_{\mathrm{dev}}$) was 5.5 points.
The last observed version matched the observed peak in 46 builds ($\Delta_{\mathrm{drop}}=0$) and fell at least one point below it in 20.
In all, 72 of the 77 builds (93.5\%) improved on their first score by at least one point at some stage, and the median time to the first peak was 4.0 hours into the six-hour budget.
Figure~\ref{fig:iteration-gain} tracks changes in the evidence returned on a fixed set of \texttt{self\_check} questions within each build, using Qwen3.5-4B to answer.
We exclude checks with fewer than 50 questions, invalid predictions, or zero scores.
Code changes can leave evidence unchanged, and the last eligible check can precede the frozen submission.
The drop from the star to the endpoint in Figure~\ref{fig:iteration-gain} measures how much of the peak development score was lost by that last check.
We use the frozen submission's hidden-test score to assess generalization.

GPT-5.6 Luna had the largest median net gain on its own development questions (10.0 points), yet it ranked tenth on the hidden tests (Table~\ref{tab:main-results}).
GPT-6.1 Sol ranked first with a smaller median gain of 4.8 points.
Luna ran 85\% of its \texttt{self\_check} commands on the full development file, compared with at most 29\% for any other configuration.
Around hour three on NQ, Luna scored 53.0 on 100 questions it had tested repeatedly and 35.0 on 100 previously unchecked questions in the same check.
Across the 27 builds with a check after hour three covering at least 30 questions in each group, the median gap favoring previously checked questions was 2.3 points.
For 20 of these builds, other configurations had tested the same questions.
We grouped these questions by whether the original agent had already tested them, using the same two groups when scoring the other configurations.
Under this grouping, the other configurations also scored 3.3 points higher on the original agent's previously checked questions than on its new questions.
The same advantage across configurations suggests that question difficulty may explain part of the gap.
Agents may reuse training examples. Differences in question selection and check eligibility limit comparisons of learning efficiency across builds.

\subsection{Development Resource Allocation}
\label{sec:resource-allocation}

\begin{figure}[t]
    \centering
    \includegraphics[width=\columnwidth]{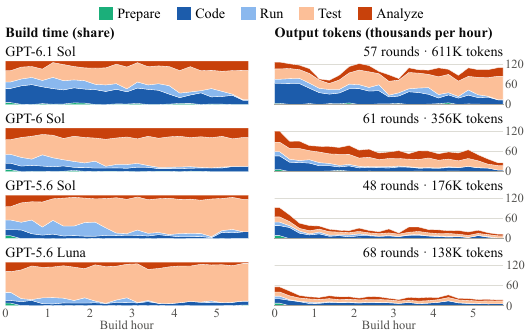}
    \caption{\textbf{Codex CLI development activity.} Panels show each activity’s share of build time and output tokens per hour, averaged across nine tasks.}
    \label{fig:resource-allocation}
\end{figure}

For the 36 Codex CLI builds, we assign elapsed time (including waits) and output tokens to Prepare, Code, Run, Test, or Analyze, counting overlapping time once.
Code and Run together cover implementation (Section~\ref{sec:process-evaluation}).
The rules agree with model labels on 50/58 calls sampled by time and 49/53 sampled by tokens. The labeling model helped develop the rules, and independent human validation remains open.

Testing and analysis took most of the development budget in these builds. Averaged over nine tasks, these activities accounted for 57--87\% of build time and 55--70\% of output tokens, as Figure~\ref{fig:resource-allocation} shows. GPT-6 Sol, GPT-5.6 Sol, and GPT-5.6 Luna spent 60--77\% of their time on Test, which accounted for 26--41\% of their tokens. Coding took under 10\% of their time and accounted for 22--33\% of their tokens. Test time includes waits for evaluation jobs, so its share also depends on how long those jobs block the agent. GPT-6.1 Sol launched the most evaluation-related jobs by our rule-based count, a median of 148 per build, yet spent an average of 0.27 hours per build waiting for them. GPT-5.6 Sol and GPT-5.6 Luna spent 2.66 and 3.27 hours, respectively.

All four configurations spent a larger share of time testing in the last hour than in the first, but agents continued to revise their systems. Code accounted for 19--32\% of final-hour output tokens, and 33 of 36 builds included at least one round of modification and evaluation every hour. We count a round when an agent edits or rebuilds the system and then launches an evaluation, as in the HotpotQA trace in Figure~\ref{fig:build-case}. The median ranged from 48 to 68 rounds per build. GPT-5.6 Luna completed the most rounds and ranked tenth by mean score, while GPT-5.6 Sol completed the fewest and ranked third. In thousands of tokens, median output per round was 11.0 for GPT-6.1 Sol, 5.8 for GPT-6 Sol, 3.9 for GPT-5.6 Sol, and 2.5 for GPT-5.6 Luna. This order matches the mean score ranking in Table~\ref{tab:main-results}. This comparison cannot separate output volume from model capability. Across the 36 builds, the correlation between output per round and score was $r=-0.40$ after we removed task and configuration means.

\subsection{Improving Human-Designed Methods}
\label{sec:improving-human-sota}

\begin{figure}[t]
    \centering
    \includegraphics[width=\columnwidth]{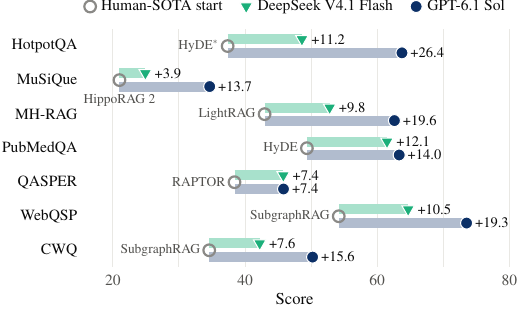}
    \caption{\textbf{Agent gains over human-designed systems.} Labels give score gains averaged over five answering models after six hours.}
    \label{fig:human-sota-improvement}
\end{figure}

For each of seven tasks, we select a Human-SOTA method suited to the task with a reusable implementation.
We use HippoRAG~2 for MuSiQue, LightRAG for MH-RAG, RAPTOR for QASPER, SubgraphRAG for WebQSP and CWQ, and HyDE for PubMedQA and HotpotQA.
GPT-6.1 Sol (Codex CLI) and DeepSeek V4.1 Flash (DeepSeek Harness) each receive the code, dependencies, index, and retrieval interface and have six hours to improve the system while preserving its core mechanism.
We compare the starting and frozen systems on hidden questions with the five answering models (Section~\ref{sec:outcome-evaluation}).

Figure~\ref{fig:human-sota-improvement} shows gains for all 14 reported builds.
All pass the integrity audit.
GPT-6.1 Sol gains 7.4 to 26.4 points, with a mean of 16.6.
DeepSeek gains 3.9 to 12.1 points, with a mean of 8.9.
GPT-6.1 Sol scores higher on all seven tasks.
The two agents nearly tie on QASPER, at 45.82 and 45.79.
In a separate repeat on CWQ, the score fell from 34.58 to 33.20.

On HotpotQA, both agents extend HyDE's hypothetical-document retrieval~\cite{gao-etal-2023-precise}.
HyDE averages the Contriever embeddings of the question and hypothetical documents to retrieve original passages by exact inner-product search.
DeepSeek adds two rounds of hypothetical-document generation from retrieved passages, adds candidates by title matching, and trains a cross-encoder on the supplied examples to select the top 30 passages.

GPT-6.1 Sol finds that second-hop passages can rank below weaker candidates.
It also observes incorrect answers on questions for which retrieval finds both supporting documents.
After an additional retrieval round, a reranker uses first-hop evidence to score candidates, and linked titles expand the search to missing bridge articles.
Qwen3.5-4B selects the supporting passages from this expanded pool.
From HyDE's hidden-test score of 37.37, DeepSeek reaches 48.61 and GPT-6.1 Sol reaches 63.74.
In a paired check on 200 validation questions, title expansion raises Qwen3.5-4B answer F1 from 77.58 to 78.86.

On MuSiQue, GPT-6.1 Sol and DeepSeek raise HippoRAG~2's score from 21.00 to 34.65 and 24.92, exceeding their respective builds from scratch by 1.74 and 3.44 points (Table~\ref{tab:main-results}).
Across seven tasks, their adapted systems average 1.92 and 1.36 points below their builds from scratch.
The adapted builds start with working code and a prebuilt index, but the agents must retain the original retrieval mechanism.
Agents building from scratch can choose that mechanism.
Estimating the effect of the starting assets alone would require the same design constraints in both settings.

\subsection{Comparison with Automated RAG Frameworks}
\label{sec:framework-comparison}

\begin{figure}[t]
    \centering
    \includegraphics[width=\columnwidth]{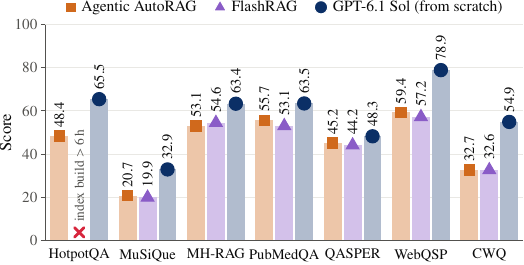}
    \caption{\textbf{Automated RAG frameworks against GPT-6.1 Sol builds from scratch.} Bars show hidden-test scores. FlashRAG did not finish indexing HotpotQA within six hours.}
    \label{fig:framework-comparison}
\end{figure}

Agentic AutoRAG~\cite{strand2026agentic} uses an LLM to diagnose errors and propose pipeline configurations.
Our FlashRAG~\cite{jin2025flashrag} baseline selects among ten predefined retrieval pipelines.
We run both on the same seven tasks with six-hour build budgets.
They select configurations on 200 training questions using Qwen3.5-4B and the task metric.
AutoRAG uses GPT-6.1 Sol as its optimizer, matching the agent in Table~\ref{tab:main-results}.
We freeze the selected systems and evaluate answer accuracy with the five answering models.

Figure~\ref{fig:framework-comparison} compares the selected systems with the agent's builds from scratch.
The agent outperforms every delivered framework system.
On the six tasks both frameworks complete, AutoRAG averages 44.5 and FlashRAG averages 43.6.
The agent averages 57.0, and Reference RAG averages 45.0.
The gaps are smallest on QASPER, at 3.1 and 4.1 points.
They reach 19.6 to 22.3 points on the two knowledge-graph tasks.
On the four common text tasks, the agent leads by 8.3 and 9.1 points on average.
Our adapter omits some retrieval and generation diagnostics or leaves them at their defaults, limiting how much of the score gap we can attribute to the available actions.

All 18 agent-built knowledge-graph systems use graph traversal (Section~\ref{sec:final-strategies}), an option absent from both framework search spaces.
On text tasks, the HotpotQA case illustrates how an agent can extend an existing pipeline with a new reranker and evidence selector.

Indexing time and token limits further restrict the search.
AutoRAG evaluates 18--60 configurations on six tasks and four on HotpotQA.
FlashRAG tests six or seven candidates on five tasks before hitting a per-question token limit and skips three candidates on MH-RAG that exceed the context limit.
All six pipelines it selects use single-pass retrieval.
On HotpotQA (5.23 million records), FlashRAG spends about three hours on the first dense index and misses the deadline on the second.
These results suggest that specialized builders could reuse indexes across trials and let agents implement missing retrieval components.
\section{Related Work}
In this section, we review related work on RAG methods and automated RAG optimization.

\textbf{RAG methods.}
RAG methods differ in how they organize and retrieve evidence.
Passage-based methods~\cite{lewis2020retrieval,izacard2021leveraging,Karpukhin2020DensePR,zhao2026qchunker,mo2026opendecoder,wei2025instructrag} retrieve relevant text passages and use them to generate answers.
Hierarchical methods~\cite{sarthi2024raptor,10.14778/3819518.3819556,zheng2026should} organize evidence at multiple levels. For example, RAPTOR~\cite{sarthi2024raptor} builds trees of text summaries, while BookRAG~\cite{10.14778/3819518.3819556} combines document structure with entity relations.
Graph-based methods~\cite{Zhuang2025LinearRAGLG,Su2026ClueRAGTA,pmlr-v267-gutierrez25a,edge2024local,Guo2024LightRAGSA,li2024subgraphrag} organize entities and their relationships to find evidence across passages.
Clue-RAG~\cite{Su2026ClueRAGTA} connects text chunks, knowledge units, and entities in a multi-partite graph for iterative retrieval.
Methods for query reformulation and retrieval control~\cite{wang2026chain,guan2026deeprag,gao-etal-2023-precise,asai2024self,yan-etal-2026-corrective,Jeong2024AdaptiveRAGLT} complement these indexing choices.
CoRAG~\cite{wang2026chain} reformulates queries based on previous retrieval results, while DeepRAG~\cite{guan2026deeprag} breaks questions into steps and decides at each step whether to retrieve external knowledge.
These studies propose RAG methods and evaluate their retrieval or answer quality.
In contrast, AutoRAGBench evaluates whether LLM agents can build and improve RAG systems for different tasks.

\textbf{Automated RAG optimization.}
Existing work on automated RAG optimization includes search-based methods and LLM-assisted methods.
Search-based methods~\cite{kim2024autorag,fu2024autorag,orbach2025raghpo,pmlr-v293-conway25a,11629401} select components, tune parameters, and optimize retrieval strategies based on evaluation results. AutoRAG~\cite{kim2024autorag} selects components one pipeline stage at a time, while AutoRAG-HP~\cite{fu2024autorag} uses hierarchical multi-armed bandits to adjust hyperparameters based on evaluation results. RAG-HPO~\cite{orbach2025raghpo} compares greedy, random, and Bayesian search, and syftr~\cite{pmlr-v293-conway25a} uses multi-objective Bayesian optimization to balance accuracy and cost in RAG pipelines with and without agents.
CARROT~\cite{11629401} uses Monte Carlo tree search (MCTS) to optimize chunk selection and ordering under a cost budget.
LLM-assisted methods analyze experimental results to guide configuration choices. HAMMER~\cite{zhou2026hammer} uses an LLM to analyze previous configurations and query results and estimate the performance of new configurations. It stores the resulting summaries in a hierarchical graph memory that guides MCTS. Agentic AutoRAG~\cite{strand2026agentic} uses LLM agents to identify whether failures occur during retrieval or generation and suggest the next configuration within a user-defined search space. Its accuracy experiments evaluate the selected configurations on separate test questions. These optimization methods operate within predefined configuration spaces.
In contrast, AutoRAGBench evaluates whether LLM agents can build and improve RAG systems from scratch, without restricting them to a predefined configuration space or prescribed design strategies.

\section{Conclusion}
In this paper, we proposed AutoRAGBench, the first benchmark to evaluate how well LLM agents develop RAG systems from scratch in an open code space. We evaluated nine agent configurations across nine RAG tasks using five answering models. The RAG systems developed by the best-performing agent configuration achieved an average score of 56.1, demonstrating that LLM agents could develop RAG systems with promising performance. We released AutoRAGBench and its evaluation resources to support further research on evaluating and improving autonomous RAG development. As future work, we plan to incorporate additional RAG benchmarks, evaluate more advanced LLM agents, and explore broader RAG development scenarios involving multimodal retrieval.

\bibliographystyle{IEEEtran}
\bibliography{refs}

\end{document}